%% file: iclr2020_conference.tex
\documentclass{article} % For LaTeX2e
\usepackage{iclr2020_conference,times}

\input{math_commands.tex}

\usepackage{hyperref}
\usepackage{url}
\usepackage{graphicx}
\usepackage{enumitem}
\usepackage{algorithm,algorithmicx}
\usepackage[noend]{algpseudocode}
\algrenewcommand\algorithmicrequire{\textbf{Input:}}
\usepackage{booktabs}
\usepackage{multirow}
\usepackage[]{todonotes}
\usepackage{mathtools}
\usepackage{pgfplots}
\usepackage{listings}
\usepackage{pgfplots}
\usepackage{pgfplotstable}
\usepackage{soul}
\usepackage{subcaption}
\usepgfplotslibrary{groupplots}
\pgfplotsset{compat=1.3}

\newcommand{\note}[4][]{\todo[author=#2,color=#3,size=\scriptsize,fancyline,caption={},#1]{#4}} % default note settings, used by macros below

\newcommand{\philipp}[2][]{\note[#1]{philipp}{blue!40}{#2}}

\newcommand{\Philipp}[2][]{\philipp[inline,#1]{#2}\noindent}

\author{
  Xutai Ma$^{1,2}$, Juan Pino$^1$, James Cross$^1$, Liezl Puzon$^1$, Jiatao Gu$^1$ \vspace*{0.2cm} \\
$^1$Facebook\\
$^2$Johns Hopkins University \vspace*{0.2cm} \\
  {\tt xutai\_\thinspace ma@jhu.edu} \\
  {\tt \{juancarabina,jcross,lie,jgu\}@fb.com}\\ }

\graphicspath{ {figures/} }
\iclrfinalcopy % Uncomment for camera-ready version, but NOT for submission.
\begin{document}

\title{Monotonic Multihead Attention}
% \xutai{How about Monotonic Multiple or Multiplex Attention, it implies both multi layer and multi head.}
\maketitle
\input{sections/abstract}
\input{sections/introduction}
\input{sections/model}

\input{sections/experiment}
\input{sections/result}

\input{sections/related-work.tex}
\input{sections/conclusion.tex}

\bibliography{iclr2020_conference}
\bibliographystyle{iclr2020_conference}

\appendix
\input{sections/appendix}

\end{document}

%% file: math_commands.tex
\usepackage{amsmath,amsfonts,bm}

\def\eqref#1{equation~\ref{#1}}
\def\1{\bm{1}}

\def\vg{{\bm{g}}}
\def\vh{{\bm{h}}}

\def\vx{{\bm{x}}}
\def\vy{{\bm{y}}}

\DeclareMathAlphabet{\mathsfit}{\encodingdefault}{\sfdefault}{m}{sl}
\SetMathAlphabet{\mathsfit}{bold}{\encodingdefault}{\sfdefault}{bx}{n}

%% file: sections/abstract.tex
\begin{abstract}
Simultaneous machine translation models start generating a target sequence before they have encoded or read the source sequence. 
Recent approaches for this task either apply a fixed policy on a state-of-the art Transformer model, 
or a learnable monotonic attention on a weaker recurrent neural network-based structure.
In this paper, we propose a new attention mechanism, Monotonic Multihead Attention (MMA), 
which extends the monotonic attention mechanism to multihead attention.
We also introduce two novel and interpretable approaches for latency control that are specifically designed for multiple attentions heads.
We apply MMA to the simultaneous machine translation task and demonstrate better latency-quality tradeoffs compared to MILk, the previous state-of-the-art approach. We also analyze how the latency controls affect the attention span and we
motivate the introduction of our model by analyzing the effect of the number of decoder layers and heads on quality and
latency.
% Code will be released upon publication.
% \liezl{Can we say the code will be released upon publication?}
\end{abstract}

%% file: sections/introduction.tex
\section{Introduction}
Simultaneous machine translation adds the capability of a live interpreter to machine translation: a simultaneous machine translation model starts generating a translation before it has finished reading the entire source sentence. Such models are useful in any situation where translation needs to be done in real time. For example, simultaneous models can translate live video captions or facilitate conversations between people speaking different languages. In a usual neural machine translation model, the encoder first reads the entire sentence,
and then the decoder writes the target sentence. 
On the other hand, a simultaneous neural machine translation model alternates between reading the input and writing the output using either a fixed or learned policy.

Monotonic attention mechanisms fall into learned policy category. Recent work exploring monotonic attention variants for simultaneous translation include: hard monotonic attention~\citep{raffel2017online}, 
monotonic chunkwise attention (MoChA)~\citep{chiu2018mocha} 
and monotonic infinite lookback attention (MILk)~\citep{arivazhagan-etal-2019-monotonic}.
MILk in particular has shown better quality / latency trade-offs than fixed policy approaches, 
such as wait-$k$ \citep{ma-etal-2019-stacl} or wait-if-* \citep{cho2016can} policies.
MILk also outperforms hard monotonic attention and MoChA; while the other two monotonic attention mechanisms only consider a fixed reading window,
MILk computes a softmax attention over all previous encoder states, which may be the key to its improved latency-quality
tradeoffs.
These monotonic attention approaches also provide a closed form expression for the expected alignment between source and target tokens, 
and avoid unstable reinforcement learning.
% \juan{revisit this claim}

% Monotonic attention-based models present another advantage in that they are designed for simultaneous translation and
% do not rely on a reinforcement learning component~\cite{gu2017learning}.
% Another advantage of monotic attention-based models 

% TODO
% \liezl{I like the illustration of hard monotonic, MoChA, and MILk from your presentation slides. Could you add them somewhere here for a side-by-side visual comparison? This goes along with Philipp's comment. Maybe there's a good way to visualize MMA as well}
However, monotonic attention-based models, including the state-of-the-art MILk, were built on top of RNN-based models. RNN-based models have been outperformed by the recent state-of-the-art Transformer model~\citep{vaswani2017attention}, which features multiple encoder-decoder attention layers and multihead attention at each layer.

%The first one is that MIlk is not capable of streaming. 
%While the attention head moves forward, 
%the model needs to compute softmax attention over the entire source history to generate a new token.
%Although infinite lookback achieved a higher quality compared to the hard monotonic or MoChA models, 
%it is impractical to implement in real-life applications, where the length of the source may be essentially unbounded.

% \xutai{The result from all monotonic attention is not promising, we might give up the statement of streaming}
% TODO
% \xutai{Can we instead say that milk needs more calculation?}

We thus propose monotonic multihead attention (MMA), 
% along with a controllable attention span, 
which combines the strengths of multilayer multihead attention and monotonic attention. 
We propose two variants, Hard MMA (MMA-H) and Infinite Lookback MMA (MMA-IL).
MMA-H is designed with streaming systems in mind where the attention span must be limited. MMA-IL emphasizes the quality of the translation system.
We also propose two novel latency regularization methods.
The first encourages the model to be faster by directly minimizing the average latency. The second encourages the attention heads to maintain similar positions, preventing the latency from being dominated by a single or a few heads.
% \liezl{I switched the "first" and "second" method to be consistent with the order they're introduced in the Latency Control section.}
% We apply MMA to the encoder self-attention, decoder self-attention and encoder-decoder attention in the Transformer architecture.

% We proposed two three variations, hard aligned, infinite lookback and hybrid.
% We show that with different variations, we could achieve better quality-latency trade-offs.

The main contributions of this paper are:
\begin{enumerate}
    \item We introduce a novel monotonic attention mechanism, monotonic multihead attention, which enables the Transformer model to perform online decoding. This model leverages the power of the Transformer and the efficiency of monotonic attention.
    \item We demonstrate better latency-quality tradeoffs compared to the MILk model, the previous state-of-the-art, on two standard translation benchmarks, IWSLT15 English-Vietnamese (En-Vi) and WMT15 German-English (De-En).
    \item We provide analyses on how our model is able to control the attention span and we motivate the design of our model
    with an ablation study on the number of decoder layers and the number of decoder heads.
\end{enumerate}

\iffalse
\Philipp{
    you should give the main idea of the attention mechanism here, 
    including a (made-up) example how a sentence is processed and how this enables streaming processing. 
    you should also discuss how reordering is handled by the attention mechanism. 
    i presume this is still possible because you have multiple attention heads.
    }
\fi

%% file: sections/model.tex
\section{Monotonic Multihead Attention Model}
In this section, we review the monotonic attention-based approaches in RNN-based encoder-decoder models.
We then introduce the two types of Monotonic Multihead Attention (MMA) for Transformer models: MMA-H and MMA-IL.
Finally we introduce strategies to control latency and coverage.
% removed: "for simultaneous translation."
\subsection{Monotonic Attention}
\label{section:hard-monotonic-attention}
The hard monotonic attention mechanism~\citep{raffel2017online} was first introduced in order to achieve online linear time decoding 
for RNN-based encoder-decoder models.
We denote the input sequence as $\mathbf{x} = \{x_1, ..., x_T\}$, 
and the corresponding encoder states as $\mathbf{m} = \{m_1, ..., m_T\}$, 
with $T$ being the length of the source sequence. 
The model generates a target sequence $\mathbf{y} = \{y_1, ..., y_U\}$ with $U$ being the length of the target sequence.
At the $i$-th decoding step, the decoder only attends to one encoder state $m_{t_i}$ with ${t_i} = j$.
% \juan{Explanation is still unclear. TODO(Xutai): update wording since step i means we may or may not generate yi}
% \xutai{Change to "when generating"}
When generating a new target token $y_i$, 
% \juan{it's not after generating, we're deciding whether to generate yi here}
the decoder chooses whether to move one step forward or to stay at the current position based on a Bernoulli selection probability $p_{i,j}$, 
so that $t_i \ge t_{i-1}$. Denoting the decoder state at the $i$-th, starting from $j=t_{i-1}, t_{i-1} + 1, t_{i-1} + 2, ...$,
% \juan{same here, I think we're talking about decoder state si-1, not si}
this process can be calculated as follows: 
\footnote{Notice that during training, to encourage discreteness, \cite{raffel2017online} added a zero mean, unit variance pre-sigmoid noise to $e_{i,j}$.}
\begin{eqnarray}
    e_{i,j} &=& \text{MonotonicEnergy}(s_{i-1}, m_j) \\
    p_{i,j} &=& \text{Sigmoid} \left(e_{i,j}\right) \\
    z_{i,j} &\sim& \text{Bernoulli}(p_{i,j})
\end{eqnarray}
When $z_{i,j} = 1$, we set $t_{i}=j$ and start generating a target token $y_i$; otherwise, we set $t_{i}=j+1$ and repeat the process.
% \juan{maybe: start generating a target token yi}
% \juan{we set ti = j}
During training, an expected alignment $\bm{\alpha}$ is introduced in order to replace the softmax attention. 
It can be calculated in a recurrent manner, shown in \autoref{eq:monotonic_attention}: 
%
% TODO
% \liezl{I corrected the equation based on the discussion with Javad in reading group today, can you verify this is correct (k --> l)}
\begin{align}
\label{eq:monotonic_attention}
\begin{split}    
    \alpha_{i,j} &= p_{i,j} \sum_{k=1}^j \left(\alpha_{i-1,k} \prod_{l=k}^{j-1} \left(1-p_{i,l}\right) \right) \\
    &= p_{i,j}\left( (1 - p_{i,j-1})\frac{\alpha_{i,j-1}}{p_{i,j-1}} + \alpha_{i-1, j} \right)
\end{split}
\end{align}
\citet{raffel2017online} also introduce a closed-form parallel solution for the recurrence relation in \autoref{eq:monotonic_attention_parallel}:
%
% TODO
% \liezl{Are you planning to add the comment about how cumprod can be unstable because of 0's in the product and how you can deal with it by adding an epsilon? Perhaps also suggest a good value of epsilon either here or in the experimental setup section.}
\begin{equation}
\label{eq:monotonic_attention_parallel}
    \alpha_{i,:} = p_{i,:}\texttt{cumprod}(1 - p_{i,:}) \texttt{cumsum} \left(\frac{\alpha_{i-1,:}}{\texttt{cumprod}
    (1 - p_{i,:})}\right)
\end{equation}
where $\texttt{cumprod} (\vx) = [1, x_1, x_1x_2, ..., \prod_{i=1}^{|\vx| - 1}x_i]$ and  $\texttt{cumsum} (\vx) = [x_1, x_1 +x_2, ..., \sum_{i=1}^{|\vx|}x_i]$. 
In practice, the denominator in Equation \ref{eq:monotonic_attention_parallel} is clamped into a range of $(\epsilon, 1]$ 
to avoid numerical instabilities introduced by $\texttt{cumprod}$.
Although this monotonic attention mechanism achieves online linear time decoding, the decoder can only attend to one encoder state. This limitation can diminish translation quality because there may be insufficient information for reordering.
% TODO \juan{is there a justification for that?}

% TODO
% \liezl{You mentioned in reading group that attending to 1 state at a time hurts the model's reordering ability. 
% That would be a more concrete justification for why this would "affect translation quality"}
Moreover, the model lacks a mechanism to adjust latency based on different requirements at decoding time.
% \Juan{what does it mean?}
% \Xutai{Trying to say the latency it not controllable for monotonic model}
% \James{How about "Moreover, the model lacks a mechanism to adjust latency based on different requirements at decoding time"? Is the idea here?} 
To address these issues, \citet{chiu2018mocha} introduce Monotonic Chunkwise Attention (MoChA), which allows the decoder to apply softmax attention over a chunk (subsequence of encoder positions). Alternatively, \cite{arivazhagan-etal-2019-monotonic} introduce Monotonic Infinite Lookback Attention (MILk) which allows the decoder to access encoder states from the beginning of the source sequence. 
The expected attention for the MILk model is defined in \autoref{eq:milk_recurent}.
\begin{equation}
    \label{eq:milk_recurent}
   \beta_{i,j} = \sum_{k=j}^{|\vx|} \left( \frac{\alpha_{i, k} \exp(u_{i,j})}{\sum_{l=1}^k  \exp(u_{i,l})} \right)
\end{equation}
%which can be also calculated in parallel with an outer cumsum and inner reverse cumsum.
% \juan{i don't think the reader will understand this last part of the sentence}
% \xutai{I copied from MILk Paper, we can just don't put it here}

\subsection{Monotonic Multihead Attention}
\label{sec:mma}
Previous monotonic attention approaches are based on RNN encoder-decoder models with a single attention 
and haven't explored the power of the Transformer model.~\footnote{MILk was based on a strengthened RNN-based model called RNMT+. The original RNMT+ model~\citep{chen2018best} uses multihead attention, computes attention only once, and then concatenates that single attention layer to the output of each decoder layer block. However, the RNMT+ model used for MILk in \cite{arivazhagan-etal-2019-monotonic} only uses a single head.}
%
% \liezl{can we make this a footnote? It interrupts the logical flow of introducing Transformer as being the next step from the RNN-based monotonic attention approaches} 
% \xutai{I agree with Liezl. I move the RNMT+ thing to the foot note}
% \juan{MILk is based on RNMT+ which has a transformer encoder. Maybe say something like: MILk is based on RNMT+, which has a transformer encoder and an LSTM decoder, however, the attention mechanism only uses one head unlike the original RNMT+ model. And maybe let's tie this to our results on num layers/num heads to show that MHA is important.}
% \james{But RNMT+ doesn't have a transformer encoder...}
% \juan{Ah ok thanks. Then we should say something like: one of the strength of RNMT+ is MHA but MILk uses single head (and single layer attention? do double check)}
The Transformer architecture~\citep{vaswani2017attention} has recently become 
the state-of-the-art for machine translation~\citep{barrault-etal-2019-findings}.
An important feature of the Transformer is the use of a separate multihead attention module at each layer.
% (TODO: cite an ablation paper). 
Thus, we propose a new approach, Monotonic Multihead Attention (MMA), 
which combines the expressive power of multihead attention and the low latency of monotonic attention.

Multihead attention allows each decoder layer to have multiple heads, where each head can compute a different attention distribution.
Given queries $Q$, keys $K$ and values $V$, multihead attention $\text{MultiHead}(Q, K, V)$ is defined in \autoref{eq:multihead}.
\begin{equation}
\label{eq:multihead}
    \begin{aligned}
        \text{MultiHead}(Q, K, V ) &= \text{Concat}(\text{head}_1, ..., \text{head}_H)W^O \\
        \text{where} \,\,\, \text{head}_h &= \text{Attention}\left(QW_h^Q, KW_h^K, VW_h^V,\right) 
    \end{aligned}
\end{equation}
The attention function is the scaled dot-product attention, defined in \autoref{eq:scaled-dot-attention}:
\begin{equation}
    \text{Attention}(Q, K, V ) = \text{Softmax}\left(\frac{QK^T}{\sqrt{d_k}}\right)V
\label{eq:scaled-dot-attention}
\end{equation}
There are three applications of multihead attention in the Transformer model:
\begin{enumerate}
    \item The \textbf{Encoder} contains self-attention layers where all of the queries, keys and values come from previous layers.
    \item The \textbf{Decoder} contains self-attention layers that allow each position in the decoder to attend to all positions in the decoder up to and including that position.
    \item The \textbf{Decoder-encoder attention} contains multihead attention layers where queries come from the previous decoder layer and the keys and values come from the output of the encoder. Every decoder layer has an decoder-encoder attention.
\end{enumerate}
For MMA, we assign each head to operate as a separate monotonic attention in decoder-encoder attention.
% cut: "Inspired by \cite{raffel2017online},"

For a transformer with $L$ decoder layers and $H$ attention heads per layer,
we define the selection process of the $h$-th head decoder-encoder attention in the $l$-th decoder layer as
% \xutai{Should we use $s$ and $m$ here instead of $Q$, $K$}
\begin{eqnarray}
e_{i,j}^{l, h} &=&  \left(\frac{m_{j}W_{l,h}^K(s_{i-1}W_{l,h}^Q)^T}{\sqrt{d_k}}\right)_{i,j} \\
p_{i,j}^{l, h} &=& \text{Sigmoid} (e_{i,j}) \\
z_{i,j}^{l, h} &\sim& \text{Bernoulli}(p_{i,j})
\end{eqnarray}
where $W_{l, h}$ is the input projection matrix, $d_k$ is the dimension of the attention head. We make the selection process independent for each head in each layer. 
% TODO
% \juan{It would be good to tie these equations to MonotonicEnergy, Sigmoid, Bernouilli
% from equations (1), (2), (3)} 
We then investigate two types of MMA, MMA-H(ard) and MMA-IL(infinite lookback).
For MMA-H, we use \autoref{eq:monotonic_attention} in order to calculate the expected alignment for each layer each head, 
given $p_{i,j}^{l, h}$.
For MMA-IL, we calculate the softmax energy for each head as follows:
% \xutai{Same, Should we use $s$ and $m$ here instead of $Q$, $K$}
\begin{eqnarray}
u_{i,j}^{l, h} = \textrm{SoftEnergy} = \left(\frac{m_{j}\hat{W}_{l,h}^K(s_{i-1}\hat{W}_{l,h}^Q)^T}{{\sqrt{d_k}}}\right)_{i,j} 
\end{eqnarray}
and then use \autoref{eq:milk_recurent} to calculate the expected attention.
Each attention head in MMA-H hard-attends to one encoder state. On the other hand, each attention head in MMA-IL can attend to all previous encoder states. Thus, MMA-IL allows the model to leverage more information for translation, but MMA-H may be better suited for streaming systems with stricter efficiency requirements.

% \juan{i don't think asynchronous is the right word here}
At inference time, our decoding strategy is shown in \autoref{alg:mma-decoding}. 
For each $l, h$, at decoding step $i$, 
we apply the sampling processes discussed in \autoref{section:hard-monotonic-attention} individually and set the encoder step at $t^{l,h}_i$. 
Then a hard alignment or partial softmax attention from encoder states, shown in Equation \ref{equation:context-vector}, 
will be retrieved to feed into the decoder to generate the $i$-th token.
The model will write a new target token only after all the attentions have decided to write -- the heads that have decided to write must wait until the others have finished reading.
%\juan{second part of the sentence seems redundant with the first part}
%\liezl{I reworded it, I think the second part adds a bit more to understanding, but feel free to delete this first if we're over the limit }
%
\begin{align}
\label{equation:context-vector}
\begin{split}
    c^l_i &= \text{Concat}(c^{l,1}_i, c^{l,2}_i, ..., c^{l,H}_i) \\
    \text{where } c^{l, h}_i &= f_{\text{context}}(\vh, t^{l,h}_i) = \begin{dcases}
    m_{t^{l,h}_i} & \text{MMA-H}\\
    \frac{ \sum_{j=1}^{t^{l,h}_i}  \exp\left(u_{i,j}^{l, h} \right) m_{j}}{\sum_{j=1}^{t^{l,h}_i}  \text{exp}\left(u_{i,j}^{l, h}\right)} & \text{MMA-IL}
    \end{dcases}
\end{split}
\end{align}

\begin{figure}
    \centering
    \includegraphics[width=0.9 \textwidth]{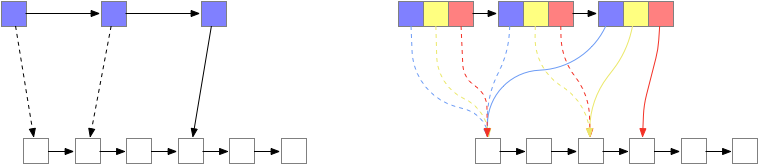}
    \caption{Monotonic Attention (Left) versus Monotonic Multihead Attention (Right). At each decoding step, MMA still has access to various encoder states.}
    % TODO: extend caption to be more informative
    \label{fig:attentions}
\end{figure}

Figure \ref{fig:attentions} illustrates a comparison between our model and the  monotonic model with one attention head.
Compared with the monotonic model, the MMA model is able to set attention to different positions so that it can still attend to previous states while reading each new token. Each head can adjust its speed on-the-fly.
% cut: "as it moves forward"
Some heads read new inputs, while the others can stay in the past to retain the source history information.
Even with the hard alignment variant (MMA-H), 
the model is still able to preserve the history information by setting heads to past states.
In contrast, the hard monotonic model, which only has one head, loses the previous information at the attention layer.
% TODO
% \xutai{The description here is not very clear here. The argument I tried to make is that the althouhg MMA-H is hard aligned, but with multiple attentions
% it can have adequete information for decoder.}
% \james{still need figure}

\begin{algorithm}
  \caption{MMA monotonic decoding. Because each head is independent, we compute line 3 to 10 in parallel}
  \label{alg:mma-decoding}
  \begin{algorithmic}[1]
    \Require{Memory $h$ of length $T$, $i=1, j=1, t^{l,h}_0=1, y_0=\text{StartOfSequence}$.}
    \While{$y_{i-1} \neq \text{EndOfSequence}$}
        \For{$l \gets 1 \textrm{ to } L$}
            \For{$h \gets 1 \textrm{ to } H$}
                \For{$j \gets t_{i-1}^{l,h} \textrm{ to } T$}
                    \State $p_{i,j}^{l,h} = \text{Sigmoid}\left(\text{MonotonicEnergy}(s_{i-1, m_j})\right)$
                    \If{$p_{i,j}^{l,h} > 0.5$}
                        \State $c^{l,h}_i = f_{\text{context}}(\vh, t^{l,h}_i) $
                        \State  $t_{i}^{l,h} = j$
                        \State \textbf{Break}
                    \EndIf
                \EndFor
            \State $c^{l}_i = \text{Concat}(c^{l,1}_i, c^{l,2}_i, ..., c^{l,H}_i)$
            \EndFor
        \State $s^l_i = \text{DecoderLayer}^l(s^l_{1:i-1}, s^{l-1}_{1:i-1}, c^l_i)$
        \EndFor
    \State $y_i = \text{Output} (s^L_i)$
    \State $i = i+1$
    \EndWhile
  \end{algorithmic}
\end{algorithm}
%\juan{In the algorithm, h designates encoder state and also head index}
% \xutai{Please have a look at the algorith, I think " we can compute line 3 to 11 in parallel"}
%\Juan{For the algorithm, consider replacing the for loop
%$\For{j \gets t_{i-1}^{l,h} \textrm{ to } T}$ by a while loop
%where the terminating condition is $z_{i,j} = 1$. 
%Finally, the break condition is based on a threshold. 
%We should probably say somewhere that we don\'t sample and use thresholding instead}

\subsection{Latency Control}
\label{sec:latency-control}
% TODO
% \xutai{Do we need some figures here to explain the idea}
Effective simultaneous machine translation must balance quality and latency. 
At a high level, latency measures how many source tokens the model has to read until a translation is generated.
The model we have introduced in \autoref{sec:mma} is not able to control latency on its own.
While MMA allows simultaneous translation by having a read or write schedule for each head, 
the overall latency is determined by the fastest head, i.e.\ the head that reads the most.
% \liezl{should be "determined by the slowest head", not fastest, right? Or did you mean to say that the "fastest" head would be one that wants to keep reading and moving forward, meaning that the absolute "fastest" possible head is the one that reaches the end for MMA-IL. And you're saying that kind of behavior means that you're waiting to read the whole sentence? In either case, this sentence is unclear}
It is possible that a head always reads new input without producing output, which would result in the maximum possible latency.
Note that the attention behaviors in MMA-H and MMA-IL can be different. 
% TODO: explain about end state not being informative
In MMA-IL, a head reaching the end of the sentence will provide the model with maximum information about the source sentence.
On the other hand, in the case of MMA-H, reaching the end of sentence for a head only gives a hard alignment to the end-of-sentence token, which provides very little information to the decoder.
% While the reaching end of sentence gives MMA-IL the most information, % it sentences the "death" for an attention head in MMA-H.
Furthermore, it is possible that an MMA-H attention head stays at the beginning of sentence without moving forward.
% \liezl{Heads stuck at the beginning of the sentence will also affect MMA-IL, right? Have we only noticed this phenomenon in practice with MMA-H? }
% \xutai{Yes. Acually I have the observation on this.}
Such a head would not cause latency issues but would degrade the model quality since the decoder would not have any information about the input. In addition, this behavior is not suited for streaming systems.
% TODO
% \juan{we should say: a head always reads new input without writing any output. 

To address these issues, we introduce two latency control methods. 
%\xutai{TODO: explain what is $g_i$}
The first one is weighted average latency, shown in \autoref{eq:expected-delay}:
\begin{equation}
    \label{eq:expected-delay}
        g_i^{W} = \frac{\text{exp}(g_{i}^{l,h})}{\sum_{l=1}^{L} \sum_{h=1}^{H} \text{exp}(g_{i}^{l,h})} g_{i}^{l,h} 
\end{equation}
where $g_{i}^{l,h} = \sum_{j=1}^{|\vx|} j \alpha_{i, j}$.
Then we calculate the latency loss with a differentiable latency metric $\mathcal{C}$. 
% \juan{need to define $\mathcal{C}$}
% \xutai{I think we defined $\mathcal{C}$ here, do you mean more details?}
% \liezl{ I'm still confused. $\mathcal{C}$ is defined as the differentiable latency metric but I don't see it used anywhere in the equation. is  $\mathcal{C}$ equivalent to $\bm{g}^W$?}
\begin{eqnarray}
    \label{eq:L-avg}
    L_{avg} &=&  \mathcal{C} \left(\bm{g}^W\right)
\end{eqnarray}
% \juan{Need to define the terms of the equation}
Like \citet{arivazhagan-etal-2019-monotonic}, we use the Differentiable Average Lagging.
It is noticeable that, different from original latency augmented training in \cite{arivazhagan-etal-2019-monotonic}, 
\autoref{eq:L-avg} is not the expected latency metric given $\mathcal{C}$, but weighted average $\mathcal{C}$ on all the attentions. 
The real expected latency is $\hat{\vg} = \max_{l,h} \left(\vg^{l,h}\right)$ instead of $\bar{\vg}$, 
but using this directly would only affect the speed of the fastest head. \autoref{eq:L-avg}, 
however, can control every head --- the regularization has a much greater effect on the fast heads but also inhibits the slow 
heads from getting faster.
% \liezl{minor: would be nice to have some empirical evidence in the appendix to justify this point: L-avg "has a much greater effect on the fast heads but also inhibits the slow heads from getting faster". Maybe show the behavior with and without L-avg}
However, for MMA models, we found that the latency of are mainly due to outliers that skip almost every token. The weighted average latency loss is not sufficient to control the outliers.
We therefore introduce the head divergence loss, the average variance of expected delays at each step, defined in \autoref{equation:L-var}: 
% \xutai{Modified here please check}
%
\begin{eqnarray}
    \label{equation:L-var}
    L_{var} &=& \frac{1}{LH} \sum_{l=1}^{L} \sum_{h=1}^{H} \left(g_{i}^{l,h} -  \bar{g}_i\right)^2
\end{eqnarray}
where $\bar{g_i} = \frac{1}{LH}\sum{g_i}$
The final objective function is presented in \autoref{eq:objective}:
\begin{equation}
L(\theta) = -\log(\vy \mid \vx; \theta) + \lambda_{avg} L_{avg} + \lambda_{var} L_{var}
\label{eq:objective}
\end{equation}
where $\lambda_{avg}$, $\lambda_{var}$ are hyperparameters that control
both losses.
Intuitively, while $\lambda_{avg}$ controls the overall speed, $\lambda_{var}$ controls the divergence of the heads.
Combining these two losses, we are able to dynamically control the range of attention heads so that we can control the latency and the reading buffer. 
For MMA-IL model, we used both loss terms; for MMA-H we only use $L_{var}$.

%% file: sections/experiment.tex
\section{Experimental Setup}
\subsection{Evaluation Metrics}
We evaluate our model using quality and latency.
For translation quality, we use tokenized BLEU 
\footnote{We acquire the data from https://nlp.stanford.edu/projects/nmt/, which is tokenized.
We do not have the tokenizer which processed this data, thus we report tokenized BLEU for IWSLT15} for IWSLT15 En-Vi
and detokenized BLEU with SacreBLEU~\citep{post-2018-call} for WMT15 De-En.
For latency, we use three different recent metrics, 
\textbf{Average Proportion} (AP)~\citep{cho2016can}, \textbf{Average Lagging} (AL)~\citep{ma-etal-2019-stacl}
 and \textbf{Differentiable Average Lagging} (DAL)~\citep{arivazhagan-etal-2019-monotonic}~\footnote{Latency metrics are computed on BPE tokens for WMT15 De-En -- consistent with \cite{arivazhagan-etal-2019-monotonic} -- and on word tokens for IWSLT15 En-Vi. }. We remind the reader of the metric definitions in Appendix \ref{sec:latency-metrics}.

\subsection{Datasets}
\begin{table}[h]
    \centering
    \begin{minipage}{\linewidth}
        \centering
        \begin{tabular}{l c c c}
        \toprule
        Dataset   &  Train & Validation & Test\\
        \midrule
        IWSLT15 En-Vi   & 133k & 1268 & 1553 \\
        WMT15 De-En  & 4.5M & 3000 & 2169\\
        \bottomrule
        \end{tabular}
    \caption{Number of sentences in each split.}
    \label{tab:mt-datasets-statistics}
    \end{minipage}
    \begin{minipage}{\linewidth}
        \centering
        \begin{tabular}{l c c c}
        \toprule
        Dataset  &  RNN & Transformer\\
        \midrule
        IWSLT15 En-Vi   &  25.6 \footnotemark   & 28.70 \\
        WMT15 De-En  & 28.4 \citep{arivazhagan-etal-2019-monotonic} & 32.3 \\
        \bottomrule
        \end{tabular}
    \caption{Offline model performance with unidirectional encoder and greedy decoding.}
    \label{tab:mt-datasets-offline}
    \end{minipage}
    \label{tab:mt-datasets}
\end{table}
\footnotetext{ \cite{luong2015stanford} report a BLEU score of 23.0 but they didn't mention what type of BLEU score they used. This score is from our implementation on the data aquired from https://nlp.stanford.edu/projects/nmt/}
We evaluate our method on two standard machine translation datasets, IWSLT14 En-Vi and WMT15 De-En.
Statistics of the datasets can be found in \autoref{tab:mt-datasets-statistics}. 
For each dataset, we apply tokenization with the Moses \citep{koehn-etal-2007-moses} tokenizer and preserve casing. 
\begin{description}[style=unboxed,leftmargin=0cm]
\item[IWSLT15 English-Vietnamese] TED talks from IWSLT 2015 Evaluation Campaign \citep{cettolo2016iwslt}.
We follow the same settings from \cite{luong2015stanford} and \cite{raffel2017online}. 
We replace words with frequency less than 5 by \textit{$<$unk$>$}. 
%\Juan{is that true?} \Xutai{Yes I checked it before the experiments, actually I was using the exact same data from stanford mt https://nlp.stanford.edu/projects/nmt/} 
We use tst2012 as a validation set tst2013 as a test set.
\item[WMT15 German-English] We follow the setting from \cite{arivazhagan-etal-2019-monotonic}. 
We apply byte pair encoding (BPE)~\citep{sennrich2016neural} jointly on the source and target to construct a shared vocabulary 
with 32K symbols. We use newstest2013 as validation set and newstest2015 as test set.
\end{description}

\subsection{Models}
We evaluate MMA-H and MMA-IL models on both datasets.
% \juan{remove sentence above?}
% \xutai{I changed a way to say it}
The MILK model we evaluate on IWSLT15 En-Vi is based on \cite{luong2015effective} rather than RNMT+ \citep{chen2018best}.
All our offline models use unidirectional encoders: the encoder self-attention can only attend to previous states.
Offline model performance can be found in \autoref{tab:mt-datasets-offline}.
For MMA models, we replace the encoder-decoder layers with MMA and keep other hyperparameter settings the same as the offline model.  
Detailed hyperparameter settings can be found in \autoref{sec:hyperparameters}.
We use the Fairseq library~\citep{ott2019fairseq} \footnote{https://github.com/pytorch/fairseq} for our implementation.
% Code will be released upon publication.
% \xutai{Should we say we will release the code like liezl said?}

%% file: sections/result.tex
\section{Results}
% \liezl{I don't get what the blue, green, and red in the figures mean}
% \xutai{Updated. There was also MoChA as well, but I was wondering whether we need to draw MoChA because it's way worse than MILk}
% \liezl{minor: I'd prefer you add a key that has red: MILk and blue: MMA in addition to having that in the caption text. Maybe you can just have the key on the top right graph?}
% DONE

In this section, we present the main results of our model in terms
of latency-quality tradeoffs and two ablation studies. In the first one, we analyze the effect of the variance loss on the attention span. Then, we study the effect of the number of decoder layers and decoder heads on quality and latency.

\subsection{Latency-Quality Tradeoffs}
\label{sec:latency-quality}
% TODO
% \xutai{TODO: Run some high latency models on IWSLT}
We plot the quality-latency curves for MMA-H and MMA-IL in \autoref{fig:quality-latency}. 
The BLEU and latency scores on the test sets were generated by setting
a latency range and selecting the checkpoint with best BLEU score on the validation set.
% \juan{TODO(Xutai): check this is correct and update numbers.}
We use differentiable average lagging~\citep{arivazhagan-etal-2019-monotonic} when setting the latency range.
We found that for a given latency, our models obtain a better translation quality.
It is interesting to observe that even MMA-H has a better latency-quality tradeoff than MILk even though each head only attends to only one state. 
Although MMA-H is not quite yet streaming capable since both
the encoder and decoder self-attention have an infinite lookback, that model represents
a good step in that direction.
% \xutai{Add more results here}
% \xutai{Consider adjust the latency range}
\input{result/bleulatency.tex}

% \juan{we should comment why quality improves as latency decreases for MMA-H (for MMA-IL, probably a different story.}

% TODO
% \subsection{Attention Span}
% We proposed two methods in \autoref{sec:controllable-attention-span} 
% with hope that the weight average latency loss could reduce the group while the variance could eliminate the outliers. 
% Table ?? shows the sweep over the $\lambda_{avg}$ and $\lambda_{var}$ on the validation set. 

% TODO
% \subsection{Visualization}
% \xutai{TODO: An example on the best model}

\subsection{Attention Span}
In \autoref{sec:latency-control}, 
we introduced the attention variance loss to MMA-H in order to prevent outlier attention heads from
increasing the latency or increasing the attention span.
We have already evaluated the effectiveness of this method
on latency in \autoref{sec:latency-quality}.
We also want to measure the difference between the fastest and slowest heads
at each decoding step.
We define the average attention span in \autoref{eq:avg-attention-span}:
\begin{equation}
    \bar{S} = \frac{1}{|\vy|} \left(\sum_{i}^{|\vy|} \max_{l,h} t_i^{l,h} - \min_{l,h} t_i^{l,h} \right)
    \label{eq:avg-attention-span}
\end{equation}
It estimates the reading buffer we need for streaming translation.
We show the relation between the average attention span (averaged over the IWSLT and WMT test sets) versus $L_{var}$ in \autoref{fig:attention-span}. As expected, the average attention span is reduced as we increase $L_{var}$.
\input{result/attentionspan.tex}

% \juan{missing numbers}

\subsection{Effect on number of layers and number of heads}

One motivation to introduce MMA is to adapt the Transformer, which is the current state-of-the-art model for machine translation, to online decoding. Important features of the Transformer architecture include having a separate attention layer for each decoder layer block and multihead attention. In this section, we test the effect of these two components on both
the offline baseline and MMA-H from a quality and latency perspective. We report quality as measure by detokenized BLEU and latency as measured by DAL on the WMT13 validation set in \autoref{fig:decheads-declayers}. We set
$\lambda_{avg} = 0$ and $\lambda_{var} = 0.2$.
% \liezl{I don't get what each graph means. Are these 3 different translation directions?}
We can see that quality generally tends to improve with more layers and more heads for both the offline baseline and MMA-H, which motivates extending monotonic attention to the multilayer/multihead setting. We also note that latency increases accordingly. This is due to having fixed
loss weights: when more heads are involved, we should increase $\lambda_{var}$ to better control latency.

% \juan{xutai: can you review this interpretation?}

\input{result/decheaddeclayer.tex}

%% file: result/bleulatency.tex
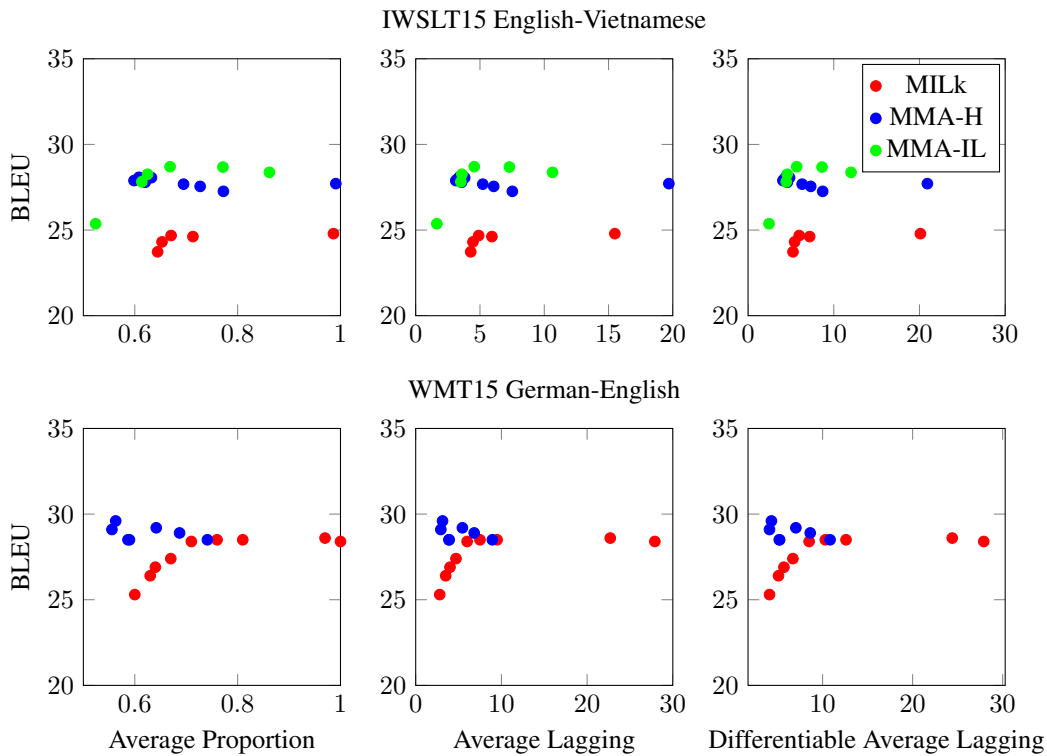
\begin{figure}
\centering
\begin{tikzpicture}
    \begin{groupplot}
    [
        group style={group size= 3 by 3, horizontal sep=1cm, vertical sep=1.5cm},
        height=5cm,
        width=5cm, 
        ]
    \nextgroupplot[ylabel=BLEU, ymin=20, ymax=35, xmin=0.5, xmax=1]
        \addplot[red, only marks] table [x index=1,y index=0, ] {result/iwslt-milk.txt};
        \addplot[blue, only marks] table [x index=1,y index=0] {result/iwslt-mma.txt};
        \addplot[green, only marks] table [x index=1,y index=0] {result/iwslt-mmail.txt};
    \nextgroupplot[title=IWSLT15 English-Vietnamese, ymin=20, ymax=35, xmin=0, xmax=20]
        \addplot[red, only marks] table [x index=2,y index=0] {result/iwslt-milk.txt};
        \addplot[blue, only marks] table [x index=2,y index=0] {result/iwslt-mma.txt};
        \addplot[green, only marks] table [x index=2,y index=0] {result/iwslt-mmail.txt};
    \nextgroupplot[ymin=20, ymax=35, xmin=0, xmax=30]
        %\addplot[red, only marks] table [x index=3,y index=0] {result/iwslt-mocha.txt};
        \addplot[red, only marks] table [x index=3,y index=0] {result/iwslt-milk.txt};
        \addlegendentry{MILk}
        \addplot[blue, only marks] table [x index=3,y index=0] {result/iwslt-mma.txt};
        \addlegendentry{MMA-H}
        \addplot[green, only marks] table [x index=3,y index=0] {result/iwslt-mmail.txt};
        \addlegendentry{MMA-IL}
    \nextgroupplot[ylabel=BLEU, ymin=20, ymax=35, xmin=0.5, xmax=1,xlabel=Average Proportion]
        %\addplot[red, only marks] table [x index=1,y index=0] {result/wmt-mocha.txt};
        \addplot[red, only marks] table [x index=1,y index=0] {result/wmt-milk.txt};
        \addplot[blue, only marks] table [x index=1,y index=0] {result/wmt-mma.txt};
    \nextgroupplot[title=WMT15 German-English, ymin=20, ymax=35, xmin=0, xmax=30, xlabel=Average Lagging]
        %\addplot[red, only marks] table [x index=2,y index=0] {result/wmt-mocha.txt};
        \addplot[red, only marks] table [x index=2,y index=0] {result/wmt-milk.txt};
        \addplot[blue, only marks] table [x index=2,y index=0] {result/wmt-mma.txt};
    \nextgroupplot[ymin=20, ymax=35 , xlabel=Differentiable Average Lagging]
        %\addplot[red, only marks] table [x index=3,y index=0] {result/wmt-mocha.txt};
        \addplot[red, only marks] table [x index=3,y index=0] {result/wmt-milk.txt};
        \addplot[blue, only marks] table [x index=3,y index=0] {result/wmt-mma.txt};
    \end{groupplot}
\end{tikzpicture}
    \caption{Latency-quality tradeoffs for MILk and MMA on IWSLT15 En-Vi and WMT15 De-En.}
    
    \label{fig:quality-latency}
\end{figure}

%% file: result/attentionspan.tex
\begin{figure}
\centering
\begin{tikzpicture}
    \begin{groupplot}
    [
        group style={group size= 2 by 1, horizontal sep=1cm, vertical sep=1.5cm},
        height=5cm,
        width=5cm, 
        ]
    \nextgroupplot[ylabel=$\bar{S}$, ymin=0, ymax=20, xmin=0, xmax=1, xlabel=$L_{var}$, title=IWSLT15 En-Vi]
        \addplot[red, mark=x] table [x index=4,y index=5] {result/iwslt-mma.txt};
     \nextgroupplot[ymin=5, ymax=20, xmin=0, xmax=2, xlabel=$L_{var}$, title=WMT15 De-En]
        \addplot[red, mark=x ] table [x index=4,y index=5] {result/wmt-mma.txt};
    \end{groupplot}
\end{tikzpicture}
    \caption{Effect of $L_{var}$ on the average attention span. The variance loss works as intended by reducing the span with higher weights.}
    \label{fig:attention-span}
\end{figure}
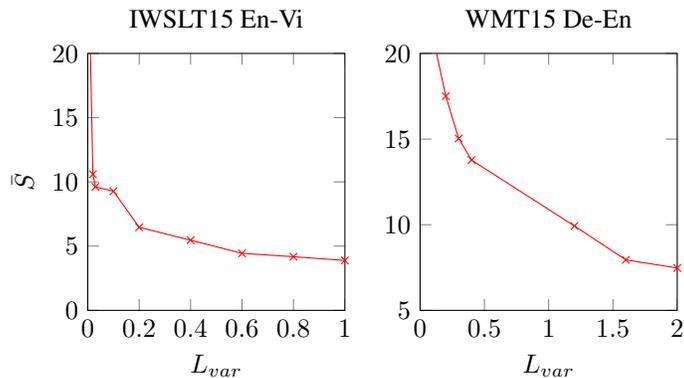

%% file: result/decheaddeclayer.tex
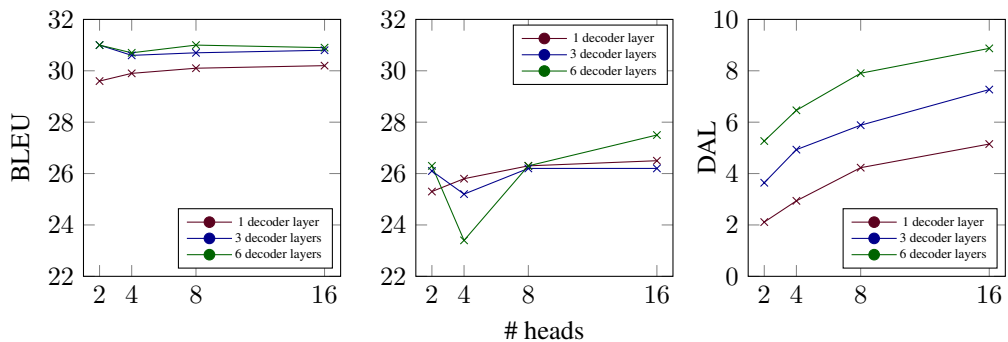
\begin{figure}
\centering
\begin{tikzpicture}
\definecolor{bordeaux}{RGB}{92,2,31}
\definecolor{darkblue}{RGB}{0,0,139}
\definecolor{darkgreen}{RGB}{0,100,0}
    \begin{groupplot}
    [
        group style={group size= 3 by 3, horizontal sep=1cm, vertical sep=1.5cm},
        height=5cm,
        width=5cm,
        ]
    \nextgroupplot[ylabel=BLEU, ymin=22, ymax=32, xmin=1, xmax=17, legend style={nodes={scale=0.5, transform shape}}, 
        legend image post style={mark=*}, xtick=data, legend pos=south east]
        \addplot[bordeaux, mark=x] table [x index=1,y index=0] {result/baseline-layer1.txt};
        \addlegendentry{1 decoder layer}
        \addplot[darkblue, mark=x] table [x index=1,y index=0] {result/baseline-layer3.txt};
        \addlegendentry{3 decoder layers}
        \addplot[darkgreen, mark=x] table [x index=1,y index=0] {result/baseline-layer6.txt};
        \addlegendentry{6 decoder layers}

    \nextgroupplot[ymin=22, ymax=32, xlabel = \# heads, xmin=1, xmax=17, legend style={nodes={scale=0.5, transform shape}}, 
        legend image post style={mark=*}, xtick=data]
        \addplot[bordeaux, mark=x] table [x index=1,y index=0] {result/mono-layer1.txt};
        \addlegendentry{1 decoder layer}
        \addplot[darkblue, mark=x] table [x index=1,y index=0] {result/mono-layer3.txt};
        \addlegendentry{3 decoder layers}
        \addplot[darkgreen, mark=x] table [x index=1,y index=0] {result/mono-layer6.txt};
        \addlegendentry{6 decoder layers}

    \nextgroupplot[ylabel=DAL, ylabel shift = -8pt, ymin=0, ymax=10, xmin=1, xmax=17, legend style={nodes={scale=0.5, transform shape}, legend pos=south east},
        legend image post style={mark=*}, xtick=data]
        \addplot[bordeaux, mark=x] table [x index=1,y index=0] {result/mono-layer1-latency.txt};
        \addlegendentry{1 decoder layer}
        \addplot[darkblue, mark=x] table [x index=1,y index=0] {result/mono-layer3-latency.txt};
        \addlegendentry{3 decoder layers}
        \addplot[darkgreen, mark=x] table [x index=1,y index=0] {result/mono-layer6-latency.txt};
        \addlegendentry{6 decoder layers}

    \end{groupplot}
\end{tikzpicture}
    
    \caption{Effect of the number of decoder attention heads and the number of decoder attention layers on quality and latency, reported on the WMT13 validation set. For both the baseline and the proposed model, quality generally improves with the number of heads and the number of layers, which motivates the proposed model.}
    \label{fig:decheads-declayers}
\end{figure}

%% file: sections/related-work.tex
% \iffalse
\section{Related Work}
% TODO
% \liezl{Add any other related work in simultaneous MT besides the 3 cited above (hard monotonic, MoChA, MILk) ?}
% \xutai{I move this to a seperate section}
% \xutai{Fahim Dalvi's paper Incremental Decoding and Training Methods for Simultaneous
% Translation in Neural Machine Translation, which is also rule based model}
Recent work on simultaneous machine translation falls into three categories.
In the first one, models use a rule-based policy for reading input and writing output.
\citet{cho2016can} propose a Wait-If-* policy to enable an offline model to decode 
simultaneously. \citet{ma-etal-2019-stacl} propose a wait-$k$ policy
where the model first 
reads $k$ tokens, then alternates reads and writes. \citet{dalvi-etal-2018-incremental} propose an incremental decoding method, also based on a rule-based schedule.
In the second category, models learn
the policy with reinforcement learning.
\cite{grissom2014don} introduce a Markov chain to phrase-based machine translation models 
for simultaneous machine translation,
in which they apply reinforcement learning to learn the read-write policy based on states.
\cite{gu2017learning} introduce an agent which learns to make decisions on when to translate from 
the interaction with a pre-trained neural machine translation model.
\cite{alinejad2018prediction} propose a new operation PREDICT which predicts future source tokens to improve quality and minimize latency.
Models from the last category leverage monotonic attention
and replace the softmax attention with an expected attention calculated from a stepwise Bernoulli selection probability.
\cite{raffel2017online} first introduce the concept of monotonic attention for online linear time decoding, 
where the attention only attends to one encoder state at a time. 
\cite{chiu2018mocha} extended that work to let the model attend to a chunk of encoder state.
\cite{arivazhagan-etal-2019-monotonic} also make use of the monotonic attention but introduce an infinite lookback to improve the translation quality.

%% file: sections/conclusion.tex
\section{Conclusion}
% In this paper, we proposed two variants of the monotonic multihead attention model for simultaneous machine translation.
% We also introduced two losses in order to control latency as well as the attention span.
% We showed that our model has better quality-latency trade-offs than the previous state-of-the-art model. Finally, we analyzed the effect of the loss we introduced
% on the attention span and we presented an ablation study on the number
% of decoder layers and heads to motivate the
% introduction of our model.

% The MMA model is not very well suited for streaming scenarios with very long source sequences since the encoder and decoder self-attention still have infinite lookback capability. In the future, we would like to investigate extensions of MMA to the encoder and decoder self-attention.

% \juan{the below is optional, maybe let's just mention streaming}
% Finally, we would also like to investigate applications of MMA-H and MMA-IL to other tasks such as speech recognition or speech translation.

In this paper, we propose two variants of the monotonic multihead attention model for simultaneous machine translation. By introducing two new targeted loss terms which allow us to control both latency 
and attention span, we are able to leverage the power of the Transformer architecture to achieve
better quality-latency trade-offs than the previous state-of-the-art model. We also present detailed 
ablation studies demonstrating the efficacy and rationale of our approach. By introducing these stronger simultaneous 
sequence-to-sequence models, we hope to facilitate important applications, such 
as high-quality real-time interpretation between human speakers.

%% file: sections/appendix.tex
\section{Appendix}
\subsection{Hyperparameters}
\label{sec:hyperparameters}
The hyperparameters we used for offline and monotonic transformer models are defined in \autoref{tab:hyperparams}.
\begin{table}[h]
    \centering
    \begin{tabular}{l c c c}
    \toprule
    Hyperparameter & WMT15 German-English & IWSLT English-Vietnamese \\
    \midrule
    encoder embed dim & 1024 & 512\\
    encoder ffn embed dim & 4096 & 1024\\
    encoder attention heads & 16 & 4\\
    encoder layers & \multicolumn{2}{c}{6} \\
    decoder embed dim & 1024 & 512\\
    decoder ffn embed dim & 4096 & 1024\\
    decoder attention heads & 16 & 4\\
    decoder layers & \multicolumn{2}{c}{6} \\
    dropout & \multicolumn{2}{c}{0.3}\\
    optimizer & \multicolumn{2}{c}{adam} \\
    adam-$\beta$ & \multicolumn{2}{c}{$(0.9, 0.98)$}  \\ 
    clip-norm & \multicolumn{2}{c}{0.0} \\
    lr & \multicolumn{2}{c}{0.0005}\\
    lr scheduler & \multicolumn{2}{c}{inverse sqrt} \\
    warmup-updates & \multicolumn{2}{c}{4000} \\
    warmup-init-lr & \multicolumn{2}{c}{1e-07} \\
    label-smoothing & \multicolumn{2}{c}{0.1} \\
    max tokens & $3584 \times 8 \times 8 \times 2$& 16000 \\
    \bottomrule
    \end{tabular}
    \caption{Offline and monotonic models hyperparameters.}
    \label{tab:hyperparams}
\end{table}

\subsection{Latency Metrics Definitions}
\label{sec:latency-metrics}

Given the delays $\textbf{g} = \{g_1, g_2, ...,g_{|\vy|}\}$ of generating each target token, AP, AL and DAL are defined in \autoref{tab:latency-mertircs}.

\begin{table}[h]
    \centering
    \begin{tabular}{l l}
    \toprule
      Latency Metric & \multicolumn{1}{c}{Calculation} \\ 
     \midrule
     Average Proportion  &  $ \displaystyle \frac{1}{|\vx||\vy|} \sum_{i=1}^{|\vy|} g_i$ \\ 
       & \\
       \midrule
      \multirow{2}{*}{Average Lagging} & $\displaystyle\frac{1}{\tau} \sum_{i=1}^{\tau} g_i - \frac{i - 1}{|\vy|/|\vx|}$  \\ 
       &  where $\tau=\arg\max_i(g_i=|\vx|)$ \\
       \midrule
      \multirow{4}{*}{Differentiable Average Lagging} 
      & $ \displaystyle \frac{1}{|\vy|} \sum_{i=1}^{|\vy|} g'_i - \frac{i - 1}{|\vy|/|\vx|}$ \\
      & where $g'_i = 
\begin{dcases}
g_i & i=0\\
\max(g_i, g'_{i-1}+\frac{|\vy|}{|\vx|}) & i< 0
\end{dcases}$\\
    \bottomrule
    \end{tabular}
    \caption{The calculation of latency metrics, given source $\vx$, target $\vy$ and delays $\vg$}
    \label{tab:latency-mertircs}
\end{table}